\documentclass[11pt,a4paper,twocolumn]{article}

\usepackage[utf8]{inputenc}
\usepackage{times}
\usepackage[colorlinks=false, pdfborder={0 0 0}]{hyperref}
\usepackage{breakurl}
\usepackage[rightcaption]{sidecap}
\usepackage{stfloats}
\usepackage{numprint}

\usepackage{multirow,bigdelim}
\usepackage{amssymb}
\usepackage{amsmath}
\usepackage{listings}
\usepackage{eurosym}
\usepackage{afterpage}

\DeclareMathOperator{\sign}{sign}

\allowdisplaybreaks

\usepackage{enumitem}

\usepackage[font=small,margin=1ex,labelfont=bf]{caption}

\usepackage{algorithm}
\usepackage{algpseudocode}

\usepackage{subfigure}

\usepackage{calc}
\usepackage[left=2.0cm,right=2.0cm,top=3.0cm,bottom=3.0cm]{geometry}

\usepackage{graphicx}

\usepackage{fancyhdr}
\fancypagestyle{firstpage}{
  \fancyhead{}
  \cfoot{\footnotesize{\em Technical Report No.~2017-03, Hochschule Niederrhein, Fachbereich Elektrotechnik \& Informatik (2017)}}
  
}

\title{\vspace*{-10mm}Paired Comparison Sentiment Scores}
\author{
Christoph Dalitz, Jens Wilberg, Katrin E. Bednarek\\
Institut f\"ur Mustererkennung\\
Hochschule Niederrhein\\
Reinarzstr. 49, 47805 Krefeld\\
{\tt christoph.dalitz{@}hsnr.de}
}
\date{}

\begin{document}

\renewcommand{\labelenumi}{\arabic{enumi})}

\twocolumn[
  \begin{@twocolumnfalse}
    \maketitle
\begin{abstract}
The method of paired comparisons is an established method in psychology. In this article, it is applied to obtain continuous sentiment scores for words from comparisons made by test persons. We created an initial lexicon with $n=199$ German words from a two-fold all-pair comparison experiment with ten different test persons. From the probabilistic models taken into account, the logistic model showed the best agreement with the results of the comparison experiment. The initial lexicon can then be used in different ways. One is to create special purpose sentiment lexica through the addition of arbitrary words that are compared with some of the initial words by test persons. A cross-validation experiment suggests that only about 18 two-fold comparisons are necessary to estimate the score of a new, yet unknown word, provided these words are selected by a modification of a method by Silverstein \& Farrell. Another application of the initial lexicon is the evaluation of automatically created corpus-based lexica. By such an evaluation, we compared the corpus-based lexica SentiWS, SenticNet, and SentiWordNet, of which SenticNet 4 performed best. This technical report is a corrected and extended version of a presentation made at the ICDM Sentire workshop in 2016.
\end{abstract}
\vspace*{2ex}

  \end{@twocolumnfalse}
  ]

\thispagestyle{firstpage}

\section{Introduction}
\label{sec:intro}
A {\em sentiment lexicon} is a dictionary that assigns each term a 
{\em polarity score}
representing the strength of the positive or negative affect associated
with the term. In general, word polarity strength depends on the context,
and its representation by a single number can therefore only be a crude
approximation. Nevertheless, such
sentiment lexica are an important tool for opinion mining and have been
proven to be very useful. Examples for recent use cases are the sentiment
analysis of tweets and SMS \cite{kiritchenko14,khiari16} or the political
classification of newspapers \cite{morik15}.

There are two approaches to building a sentiment lexicon: {\em corpus based}
automatic assignment or {\em manual annotation}. Corpus based approaches
start with a set of seed words of known polarity and extend this set with
other words occurring in a text corpus or a synonym lexicon. One possible
approach is to compute the ``Pointwise Mutual  Information'' (PMI)
\cite{church90} from co-occurrences of seed words and other words. The German
sentiment lexicon {\em SentiWS} \cite{remus10} was built in this way.
The English sentiment lexicon {\em SentiWordNet} \cite{baccianella10} is
based on the propagation of seed words, too, but by means of semi-supervised
classification and random walks. Yet another sophisticated corpus-based method
was implemented by Cambria et al.~for {\em SenticNet}
\cite{cambria14,cambria15,cambria16}.

Corpus based methods have the advantage of building large lexica in an
automated way without time consuming experiments with human annotators.
They have two drawbacks, however: due to peculiarities in the corpus, 
some words can obtain strange scores. In SentiWS 1.8, e.g., ``gelungen''
({\em successful}) has the highest positive score (1.0) while the more
positive word ``fantastisch'' ({\em fantastic})
only has a score of 0.332. In SenticNet 3.0, ``inconsequent'' has a strong
positive polarity (0.948). Moreover, it is not possible to assign a score value
to words that are absent from the corpus.

Assigning polarity scores by manual annotations can be done in two different
ways. One is by direct assignment of an ordinal score to each word on a
Likert-type scale. In this way, Wilson et al.~have created a subjectivity lexicon
with English words \cite{wilson05}, which has also been used by means of
automated translations for sentiment analysis of German texts \cite{wiegand14}.
The other method is to present words in pairs and let the observer
decide which word is more positive or more negative. Comparative studies
for other use cases have shown that scores from paired comparisons are more
accurate than direct assignments of scores \cite{mantiuk12}. The main advantage
is their invariance to scale variances between different test persons. This
is especially important when words are added at some later point when the
original test persons are no longer available.
Unfortunately, paired comparisons are much more expensive than direct
assignments:
for $n$ words, direct assignments only require $O(n)$ judgments, while a
complete comparison of all pairs requires $O(n^2)$ judgments. For large $n$,
this becomes prohibitive and must be replaced by incomplete comparisons, 
i.e.~by omitting pairs. Incomplete paired comparisons are widely deployed
in the estimation of chess players' strength \cite{elo78,batchelder79}.

In the present article, we propose a method for building a sentiment lexicon
from paired comparisons in two steps. At first, an initial lexicon is built
from a limited set of 199 words by comparison of all pairs. This lexicon is
then subsequently extended with new words, which are only compared to a limited
number of words from the initial set, which are determined by a modified version
of Silverstein \& Farrell's sorting method \cite{silverstein01}.

As we plan to use the lexicon
ourselves for the sentiment analysis of German news reporting and thus were in
need of a reliable German sentiment lexicon, and as our test persons were
native German speakers, the scope of our investigation was restricted to
German words. In order to use this lexicon for an evaluation of English
lexica, we utilized the ``averaged translation'' technique described in
Sec.~\ref{sec:results:comparison}. Automatic translation is a common method
in multilingual sentiment analysis \cite{dennecke08,korayem16}. It should
be noted, however, that the ground truth scores of our new sentiment lexicon
can only be used to compare sentiment lexica with respect to their accuracy
in the sentiment analysis of {\em German} texts.

This article is an extended and corrected version of a presentation made at
the ICDM workshop ``Sentire'' in December 2016 \cite{dalitz16}. Compared
to that presentation, the following changes and additions have been made:
\begin{enumerate}
\item The scores in \cite{dalitz16} had been computed with an approximation
  by Elo that turned out to be grossly inaccurate in this use case
  \cite{dalitz17}. The formula has been replaced by a numeric optimization
  and all scores have been recomputed. This correction also fixes the odd
  result of the leave-one-out experiment in \cite{dalitz16}.
\item The validity of the probabilistic model is checked and validated
  with a $\chi^2$ goodness-of-fit test.
\item The two large English sentiment lexica SenticNet 4.0
  \cite{cambria16} and SentiWordNet 3.0 \cite{baccianella10} have been included
  in the evaluation by means of averaged translation.
\item An error inherited from SentiWS has been corrected that had lead to a
  duplicate word in two different spellings, which has been corrected by merging
  the duplicates.
\end{enumerate}
Our article is organized as follows: Sec.~\ref{sec:method} provides
an overview over the mathematical model that underpins the method
of paired comparisons and presents formulas for computing scores from
paired comparison results,
Sec.~\ref{sec:experiments} describes the criteria for choosing the initial
set of words and our experimental setup, and Sec.~\ref{sec:results} presents
the results for the initial lexicon, evaluates the method for adding new
words, and presents an evaluation of the other lexica Sentiws,
SenticNet, and SentiWordNet by means of our ground truth initial lexicon.

\section{Method of paired comparison}
\label{sec:method}
The method of paired comparison goes back to the early 20th century
\cite{thurstone27}. See \cite{batchelder79} for a comprehensive presentation
of the model and its estimation problems, and \cite{cattelan12} for a
review of recent extensions. Applied to word polarity, it makes the assumption
that each word $w_i$ has a hidden score (or rating) $r_i$. The probability
that $w_i$ is more positive than $w_j$ (symbolically: $w_i>w_j$)
in a randomly chosen context depends on the difference between the hidden
scores:
\begin{figure}[t]
  \centering
  \includegraphics[width=0.8\columnwidth]{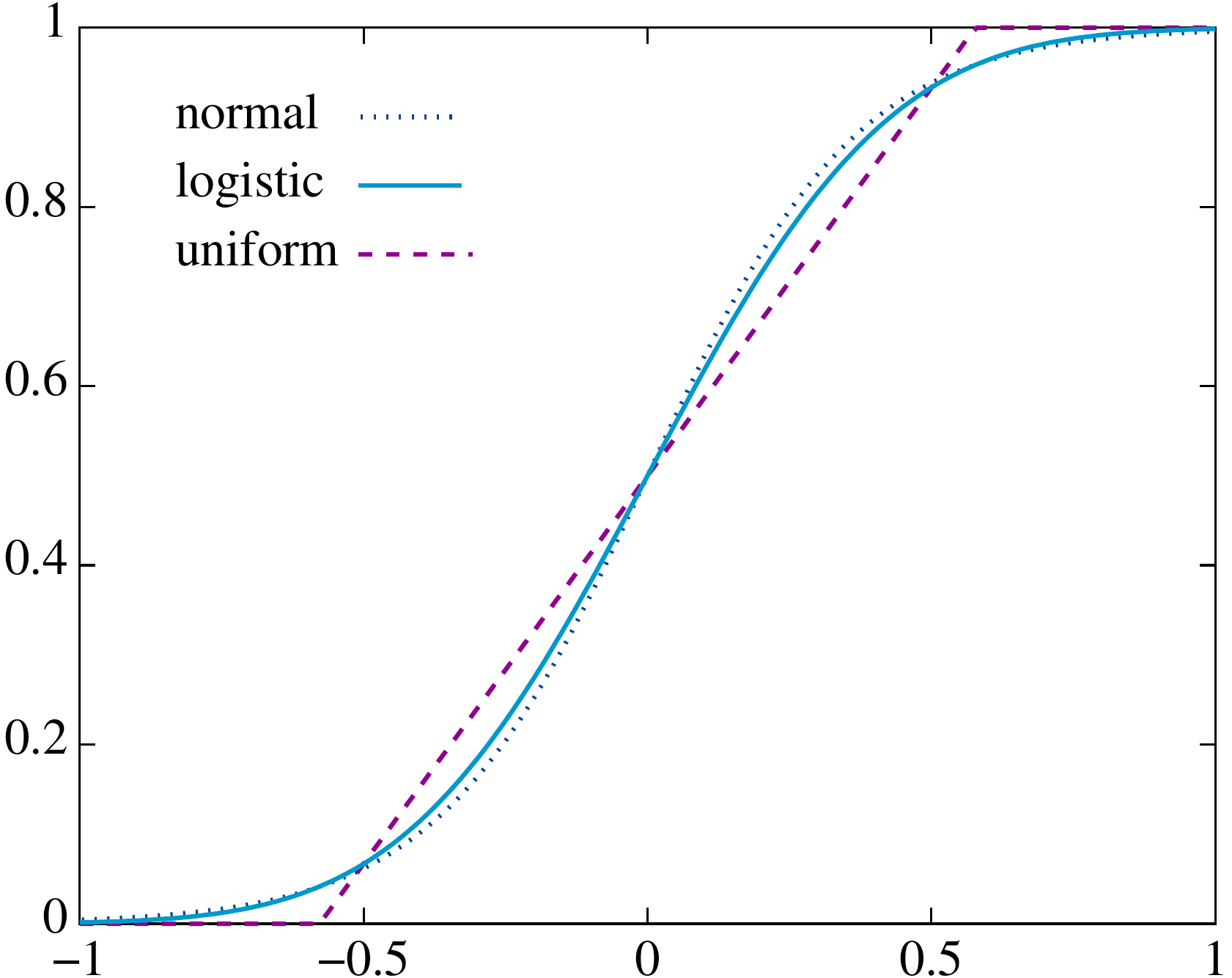}
  \caption{\label{fig:cdf} Different choices for the cumulative distribution
    function $F$ with identical standard deviations $\sigma=1/3$.}
\end{figure}
\begin{subequations}
\label{eq:ansatz}
\begin{align}
P(w_i > w_j) & = F(r_i-r_j-t) \\
P(w_i \approx w_j) & = F(r_i-r_j+t) - F(r_i-r_j-t)\\
P(w_i < w_j) & = F(r_j-r_i-t)
\end{align}
\end{subequations}
where $t$ is approximately proportional to the drawing probability of equal
strength players and can thus be considered as a {\em draw width},
and $F$ is the cumulative distribution
function of a zero-symmetric random variable. Thurstone's model \cite{thurstone27} uses an $F$
based on the normal distribution, a model that can be derived from the
assumption that the polarity of a word $w_i$ is normally distributed around its
mean inherent score $r_i$. Although this is the only model with a sound 
statistical justification, simpler distribution functions have also
been used for convenience, e.g.~the logistic distribution
(Bradley-Terry model) \cite{bradley52} or the uniform distribution
\cite{noether60}.  The shapes of these functions are quite similar
(see Fig.~\ref{fig:cdf}), and they typically provide similar fits to data
\cite{stern92}, although the asymptotic error of the score estimates 
behaves differently \cite{vojnovic16}\footnote{Beware, however, that both Stern and Vojnovic \& Yun only considered models without a draw possibility, i.e., they only considered the case $t=0$.}.
The standard deviation $\sigma$ of the distribution function $F$ is a scale
parameter that determines the range of the ratings $r_i$. We have set it to
$\sigma=1/3$ so that the estimated scores approximately fell into the range
$[-1,1]$ for our data.

As the probabilities in Eq.~(\ref{eq:ansatz}) only depend on rating differences,
the origin $r=0$ cannot be determined from the model, but must be defined
by an external constraint. Typical choices are the average rating constraint
$\sum_i r_i=0$, or the reference object constraint, i.e.~$r_i=0$ for some $i$.
For sentiment lexica, a natural constraint can be obtained by separately 
classifying words into positive and negative words and choosing the origin in
such a way that the scores from the paired comparison model coincide with
these classifications.

The ratings $r_i$ and the draw-width $t$ must be estimated from the observed
comparisons. During our two steps of building a sentiment lexicon, two different
estimation problems occur:
\begin{enumerate}
\item Estimation of one unknown $r$ of a new word from $m$ comparisons with
  old words with known ratings $q_i,\ldots,q_m$.
\item Estimation of $t$ and all unknown $r_1\ldots,r_n$ from
  arbitrary pair comparisons.
\end{enumerate}
In \cite{dalitz16}, analytic formulas were given for the estimators, but these
were based on an approximation by Elo \cite[paragraph 1.66]{elo78} which
turned out to be grossly inaccurate in this use case \cite{dalitz17}.
The formulas given in \cite{dalitz16} should therefore not be used, and
it is necessary instead to estimate the parameters via numeric optimization
either with the maximum likelihood principle or with non-linear least squares.

\subsection{Case 1: one unknown rating $r$}
\label{sec:method:case1}
Let us first consider this simpler case. 
The log-likelihood function $\ell(r)$ ($t$ is considered as given) is
\begin{align}
\label{eq:l(one_r)}
\ell(r) &= \sum_{\mbox{\scriptsize\it wins}} \log F(r-q_i-t)  \\
  &+ \sum_{\mbox{\scriptsize\it draws}} \log\Big(F(r-q_i+t) - F(r-q_i-t)\Big) \nonumber \\
  &+ \sum_{\mbox{\scriptsize\it losses}} \log F(q_i-r-t) \nonumber
\end{align}
As this is a function of only one variable $r$, it can be easily maximized
numerically, e.g., with the R function {\em optimize}.

Batchelder \& Bershed \cite{batchelder79} suggested an alternative
method-of-moments estimation by setting a
combination of the observed  number of wins $W$ and draws $D$ equal to its 
expectation value and solve the resulting equation for the parameters:
\begin{align}
\label{eq:W+D/2}
W + &D/2 = \sum_{i=1}^m F(r-q_i-t) \\
 &+ \frac{1}{2}\sum_{i=1}^m \Big(F(r-q_i+t) - F(r-q_i-t)\Big) \nonumber
\end{align}
This combination is constructed in such a way that a Taylor
expansion of the right hand side around $t=0$
cancels out the term linear in $t$, so that we obtain for small $t$
\begin{equation}
\label{eq:W+D/2approx}
W+D/2 = \sum_{i=1}^m F(r-q_i) + O(t^2)
\end{equation}
Even this approximation must be solved numerically for $r$, but it
makes the score estimation independent of the estimate for $t$, and it
shows that the draw width has only little effect on the score estimator.

It should be noted
that both the solution of (\ref{eq:W+D/2}) or (\ref{eq:W+D/2approx}) and the
maximization of (\ref{eq:l(one_r)}) yield $r=\infty$ (or $-\infty$)
when $F$ is chosen to
be the normal or logistic distribution and the word wins (or looses) all
comparisons. For the uniform distribution, the solution or minimum is not
unique in this case, but this ambiguity could be resolved by choosing the
solution with $\min(|r|)$.

\subsection{Case 2: all ratings $(r_i)_{i=1}^n$ and $t$ unknown}
\label{sec:method:case2}
The log-likelihood function in this case is
\begin{align}
\label{eq:l(all_r)}
\ell(&r_1,\ldots,r_n,t) = \sum_{\substack{\mbox{\scriptsize\it comparisons} \\ {\mbox{\scriptsize\it with }w_i>w_j}}} \!\!\log F(r_i-r_j-t)  \\
  &+ \sum_{\substack{\mbox{\scriptsize\it comparisons} \\ {\mbox{\scriptsize\it with }w_i\approx w_j}}} \!\!\log\Big(F(r_i-r_j+t) - F(r_i-r_j-t)\Big) \nonumber
\end{align}
Due to the large number of $n+1$ parameters,
numerical methods for maximizing the log-likelihood function
(\ref{eq:l(all_r)}) might be very slow or
fail to converge. In this situation, an alternative solution obtained
via the least-squares method can be helpful.

For a least squares fit of the parameters, let us consider the same
observable as in case 1, but now for each word $w_i$:
\begin{equation}
S_i = \underbrace{W_i}_{\mbox{\scriptsize wins of word $w_i$}} +\quad \frac{1}{2}\underbrace{D_i}_{\mbox{\scriptsize draws of word $w_i$}}
\end{equation}
When we set this ``scoring'' equal to its expectation value and make again a Taylor series
expansion at $t=0$, we obtain
\begin{equation}
S_i = \sum_{j\in C_i} F(r_i-r_j) + O(t^2)
\end{equation}
where $C_i$ are the indices of all words $w_j$ that have been compared with
$w_i$\footnote{Note that indices may occur more than once in $C_i$ because $w_i$ might have been compared with some other word more than once}.
Joint estimates for all ratings can then be obtained
by minimizing the sum of the squared deviations
\begin{equation}
\label{eq:ss}
SS (r_1,\ldots,r_n) = \sum_{i=1}^n\Big( S_i - \sum_{j\in C_i} F(r_i-r_j) \Big)^2
\end{equation}
Minimizing (\ref{eq:ss}) is a non-linear least squares problem, which can
be solved efficiently with the Levenberg-Marquardt algorithm as it is
provided, e.g., by the R package {\em minpack.lm} \cite{minpack.lm}.

To obtain an approximate estimator for the draw width $t$, let us consider
the total number of draws $D_i$ of each word $w_i$ as an observable and set
it equal to its expectation value from its $k_i$ comparisons:
\begin{equation}
D_i = \sum_{j\in C_i} \Big(F(r_i-r_j+t) - F(r_i-r_j-t)\Big)
\end{equation}
Keeping only the first non-zero term in a Taylor expansion around $t=0$ 
of the sum on the right hand side yields
\begin{align}
\sum_{j\in C_i} \Big(F(r_i&-r_j+t) - F(r_i-r_j-t)\Big) \nonumber\\
&\approx\quad 2t \sum_{j\in C_i} F'(r_i-r_j)
\end{align}
Again, we can determine $t$ by minimizing the sum of the squared deviations
\begin{equation}
\label{eq:ss(t)}
SS (t) = \sum_{i=1}^n \left(D_i - 2t \sum_{j\in C_i} F'(r_i-r_j) \right)^2
\end{equation}
The minimum of expression (\ref{eq:ss(t)}) can be found analytically
by solving for the zero of $SS'(t)$, which yields
\begin{equation}
\label{eq:t-lsq}
t = \frac{\sum_{i=1}^{k_i} f_iD_i/2}{\sum_{i=1}^{k_i} f_i^2} 
\quad\mbox{ with}\quad f_i = \sum_{j\in C_i} F'(r_i-r_j)
\end{equation}
The least-squares solution (\ref{eq:ss}) and (\ref{eq:t-lsq}) can either
be used as an estimator for the parameters, or it can be used as a starting
point for maximizing the log-likelihood function.

\section{Experimental design}
\label{sec:experiments}

\begin{figure*}[t]
  \centering
  \subfigure[direct assignment]{\includegraphics[height=21ex]{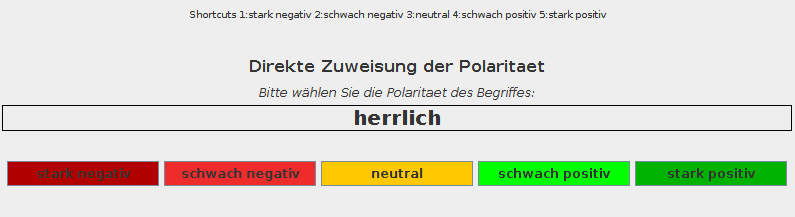}\label{fig:gui-direct}}\hspace{1em}
  \subfigure[paired comparison]{\includegraphics[height=21ex]{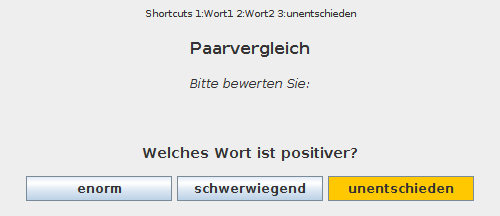}\label{fig:gui-pair}}
  \caption{\label{fig:gui} Graphical user interface for score assignment as seen by the test persons.}
\end{figure*}

To select 200 words for building the initial lexicon from round robin pair
comparisons, we have started with all $\numprint{1498}$ adjectives from SentiWS
\cite{remus10}. We have chosen adjectives because sentiments are more often
carried by adjectives and adverbs than by verbs and nouns \cite{esuli06}.
To build an intersection of these words with SenticNet
\cite{cambria14}, we translated all words into English with both of the
German-English dictionaries from {\em www.dict.cc} and {\em www.freedict.org},
and removed all words without a match in SenticNet. From the remaining
$\numprint{1303}$
words, we selected manually 10 words that appeared strongly positive to us, 
and 10 strongly negative words. This was to make sure that the polarity
range is sufficiently wide in the initial lexicon. The remaining words were
ranked by their SentiWS score and selected with equidistant ranks, such that
we obtained 200 words, with an equal number of positive and negative
words according to SentiWS. As we noticed after finishing the experiments,
there are duplicate words in different spellings in SentiWS, and we eventually
only had 199 different words by an unhappy coincidence. In the computations
of section \ref{sec:results} in the present paper, we have taken care of this
by correcting one of the spellings, i.e.~replaced the word ``phantastisch''
with ``fantastisch'' ({\em fantastic}).

We let ten different test persons assign polarity scores to these words in
two different experiments. The first one consisted of direct assignment of
scores on a five degree scale (see Fig.~\ref{fig:gui-direct}), which resulted
in ten evaluations for each word. An average score was computed for each word
by replacing the ordinal scale with a metric value ($-1$ = strong negative,
$-0.5$ = weak negative, $0$ = neutral, $0.5$ = weak positive,
$1.0$ = strong positive).

The second experiment consisted of twofold round robin paired comparisons
between the original 200 words,
with all $2\cdot\numprint{19900}$ pairs evenly distributed among the ten test persons, such that each person evaluated $\numprint{3980}$ pairs.
Due to the duplicate word, this actually was {\em not} a round robin experiment,
but one word (``fantastisch'') was compared more often than the others.
See Fig.~\ref{fig:gui-pair} for the graphical user interface presented to
the test persons.

The scores were computed both with the maximum likelihood (ML) method
and the non-linear least squares (LSQ) method as described in
section \ref{sec:method:case2}.
The standard deviation of the distribution function $F$ was set to 
$\sigma=1/3$,
which corresponds to the distribution functions in Fig.~\ref{fig:cdf}.
Maximization of the log-likelihood
function (\ref{eq:l(all_r)}) took about 12 minutes on an i7-4770 3.40 GHz
with the ``BFGS'' method of the 
R-function {\em optim}. As this method uses numerically computed gradients,
it was not applicable in the case of the uniform distribution, because its
distribution function is not differentiable. We therefore resorted to the
much slower optimization by Nelder \& Mead in this case with the Elo 
approximate solution \cite{dalitz16} as a starting point, which took 45 minutes.
The LSQ solution, on the contrary, only took 12 seconds in all cases
because we could use the more efficient Levenberg-Marquardt algorithm from the
R package {\em minpack.lm} \cite{minpack.lm} for minimizing (\ref{eq:ss}).
When the LSQ solution was
used as a starting point for the ML optimization, the runtime of the BFGS
algorithm reduced to 6 minutes. This was not applicable in the case of
the uniform distribution, though, because the LSQ solution corresponds
to a log-likelihood function that is minus infinity.

For a reasonable choice for the origin $r=0$, we shifted all scores such
that they best fitted to the discrimination between positive and negative
words from the direct comparison experiment. To be precise: 
when $r'_i$ is the score from
the direct assignment and $r_i$ the score from the paired comparisons
with an arbitrarily set origin, we chose the shift value $\rho$ that
minimized the squared error
\begin{equation}
\label{eq:error-r}
SE(\rho) \quad= \hspace{-2ex}\sum_{\sign(\rho+r_i)\neq \sign(r'_i)} \hspace{-3ex}(\rho+r_i)^2
\end{equation}

\begin{algorithm}[t]
\caption{\label{alg:nw}One-fold addition of new word}
\begin{algorithmic}[1]
\Require word $w$ with unknown rating $r$, number of comparisons $m$, words \mbox{$\vec{v}=(v_1,\ldots,v_n)$}
 sorted by their known ratings $q_1,\ldots,q_n$
\Ensure new rating $r$
\State $i_l\gets 1$ and $i_r\gets n$ 
\State $i\gets \lfloor (i_l+i_r)/2\rfloor$ \Comment{might be randomized}
\State $\vec{u}\gets \vec{v}$ and $m_0\gets 0$
\State $\vec{q}\gets ()$ and $\vec{S}\gets ()$
\While{$i>i_l$ and $i<i_r$} \Comment{binary search}
  \State $m_0\gets m_0+1$
  \State $\vec{q}\gets\vec{q}\cup q_i$
  \State $s\gets$ score from $w$ versus $v_i$ comparison,\\
  \hspace{3em}where win counts 1 and draw counts 1/2
  \State $\vec{S}\gets \vec{S}\cup s$
  \State $\vec{u}\gets\vec{u}\setminus v_i$
  \If{$s>1/2$}
    \State $i_l\gets i$
  \Else
    \State $i_r\gets i$
  \EndIf
  \State $i\gets \lfloor (i_l+i_r)/2\rfloor$ \Comment{might be randomized}
\EndWhile
\State $r_0\gets (q_{i_l} + q_{i_r})/2$ \Comment{first guess}
\State $\vec{u}\gets m-m_0$ words in $\vec{u}$ with closest ratings to $r_0$
\State $\vec{q}\gets\vec{q}$ $\cup$ ratings of $\vec{u}$
\State $\vec{S}\gets\vec{S}$ $\cup$ scores of $w$ against words from $\vec{u}$
\State $r\gets$ LSQ estimate from $(\vec{S}, \vec{q})$ \Comment{cf.~Sec.~(\ref{sec:method:case1})}
\State \Return $r$
\end{algorithmic}
\end{algorithm}

For adding new words, we implemented the method by Silverstein \& Farrell,
which uses comparison results to sort the new word into a binary
sort tree built from the initial words \cite{silverstein01}. For $n$ initial
words, this only leads to
$\log_2(n)$ comparisons, which generally are too few for computing a reliable
score. We therefore extended this method by adding comparisons with
words from the initial set which have the closest rank to the rank obtained
from the sort tree process. Algorithm \ref{alg:nw} lists the resulting 
algorithm in detail. In practice, the selection of words in the binary sort
process (lines 2 \& 18 in Algorithm \ref{alg:nw}) might be randomized by adding
a small random index offset. This would avoid that the test persons
permanently have to compare with the same pivot elements.
This algorithm can be applied sequentially to more than
one test person by estimating the resulting rating from all scores obtained
from all test persons.
We have evaluated this method with a leave-one-out experiment using
the comparisons from our two-fold round-robin comparison experiment.

\section{Results}
\label{sec:results}

\begin{figure*}
  \centering
  \subfigure[normal distribution $F$]{\label{fig:gof:norm}\includegraphics[width=0.9\columnwidth]{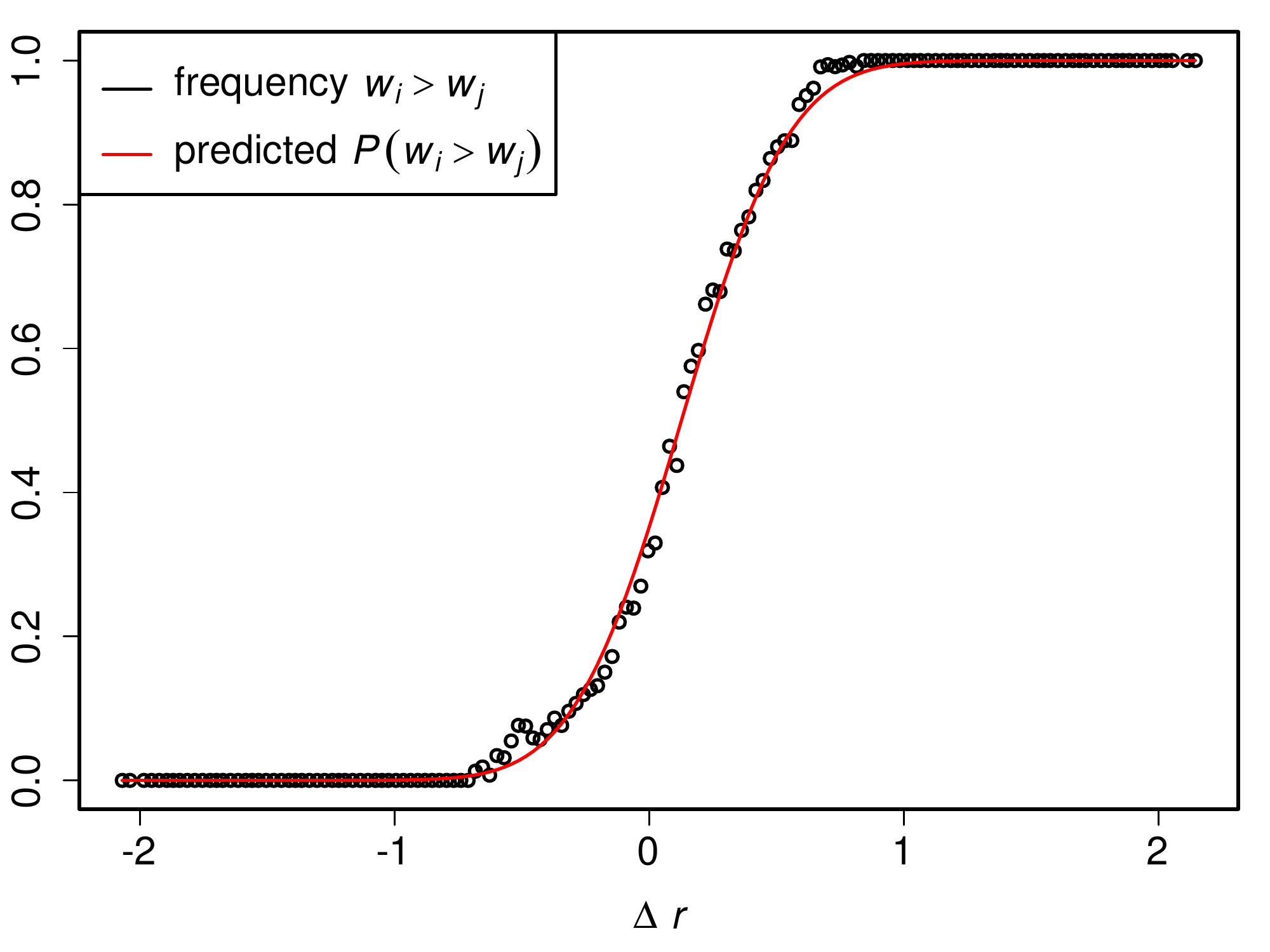}
  \includegraphics[width=0.9\columnwidth]{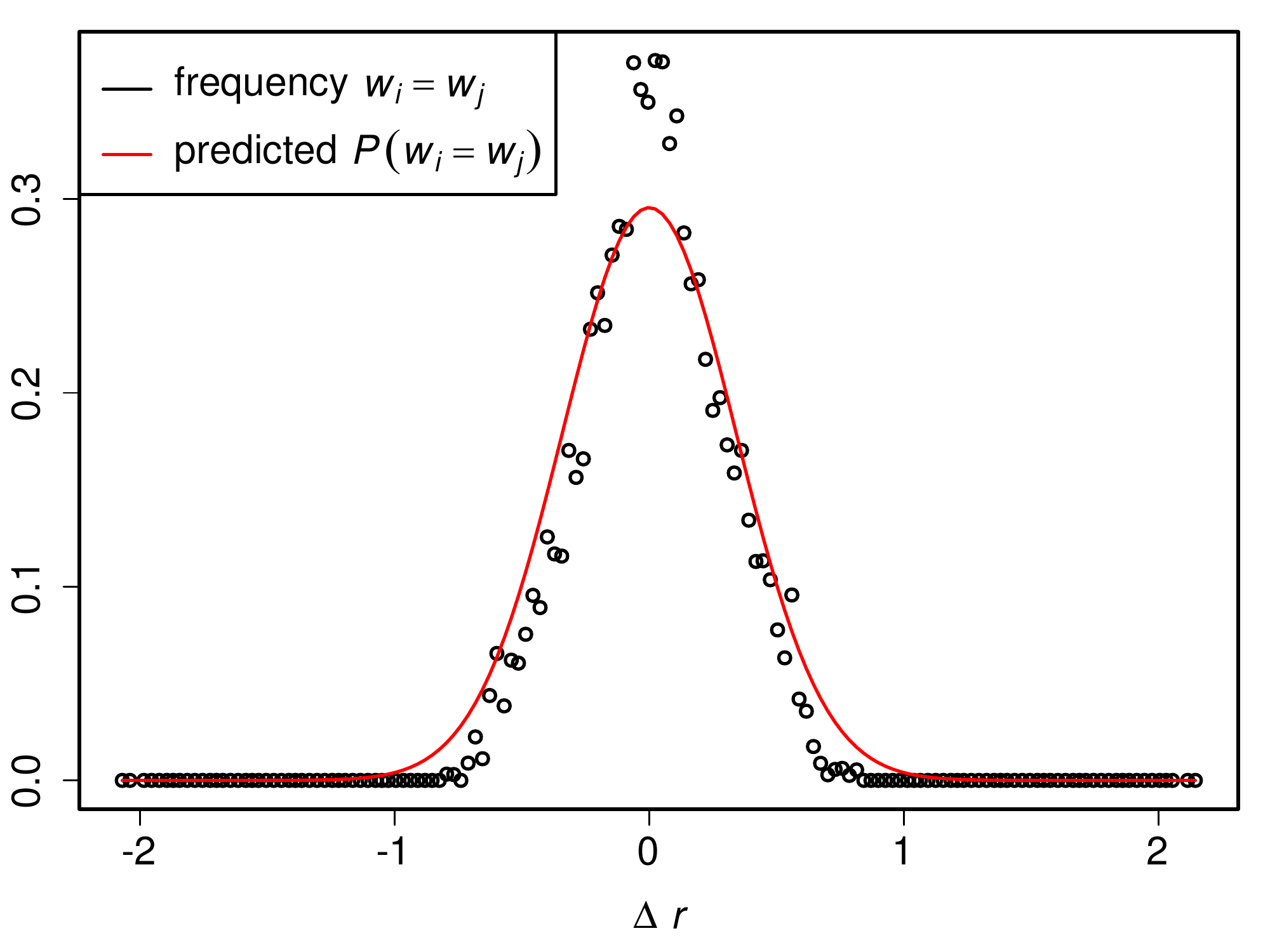}}
  \subfigure[logistic distribution $F$]{\label{fig:gof:norm}\includegraphics[width=0.9\columnwidth]{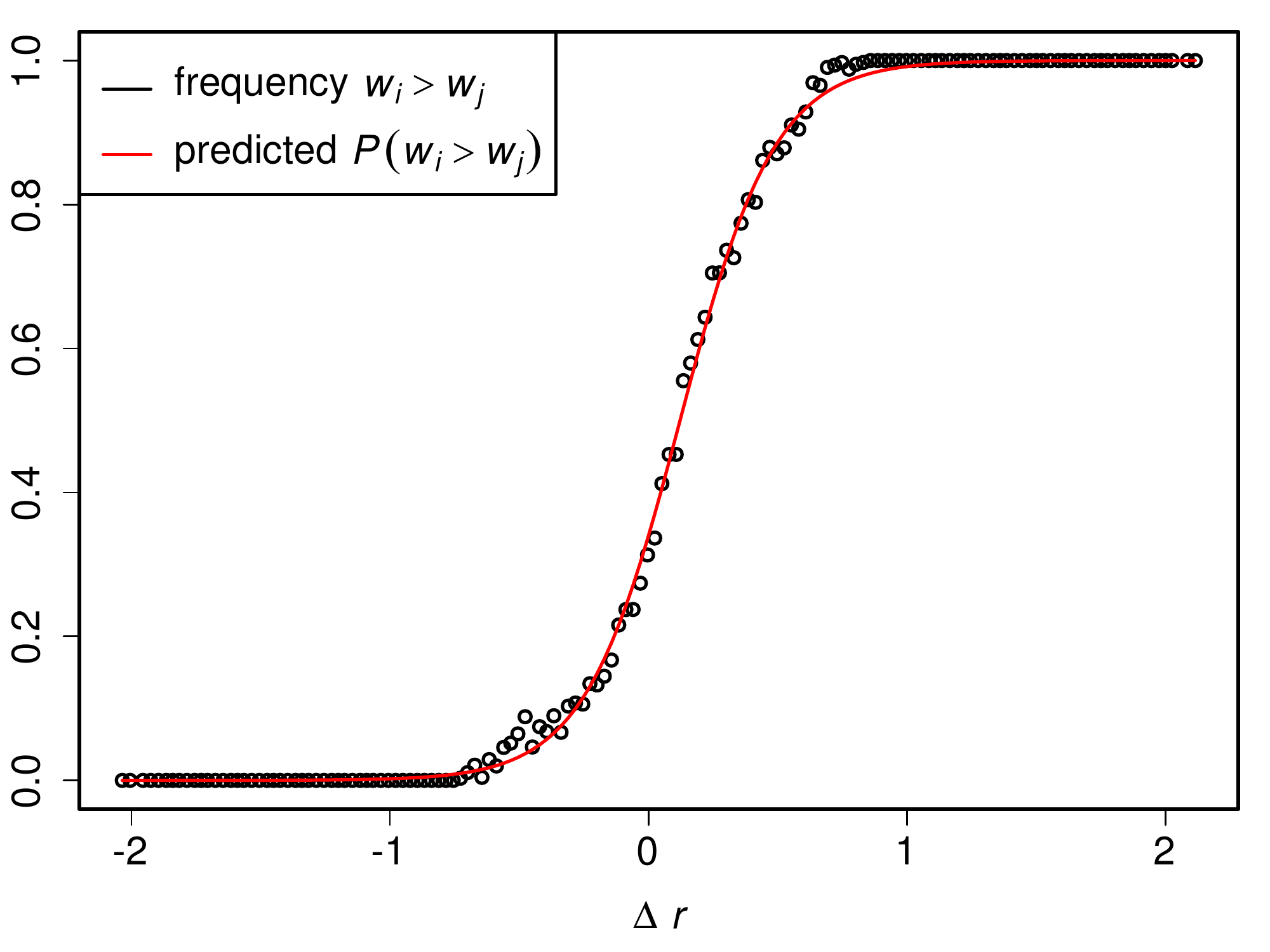}
  \includegraphics[width=0.9\columnwidth]{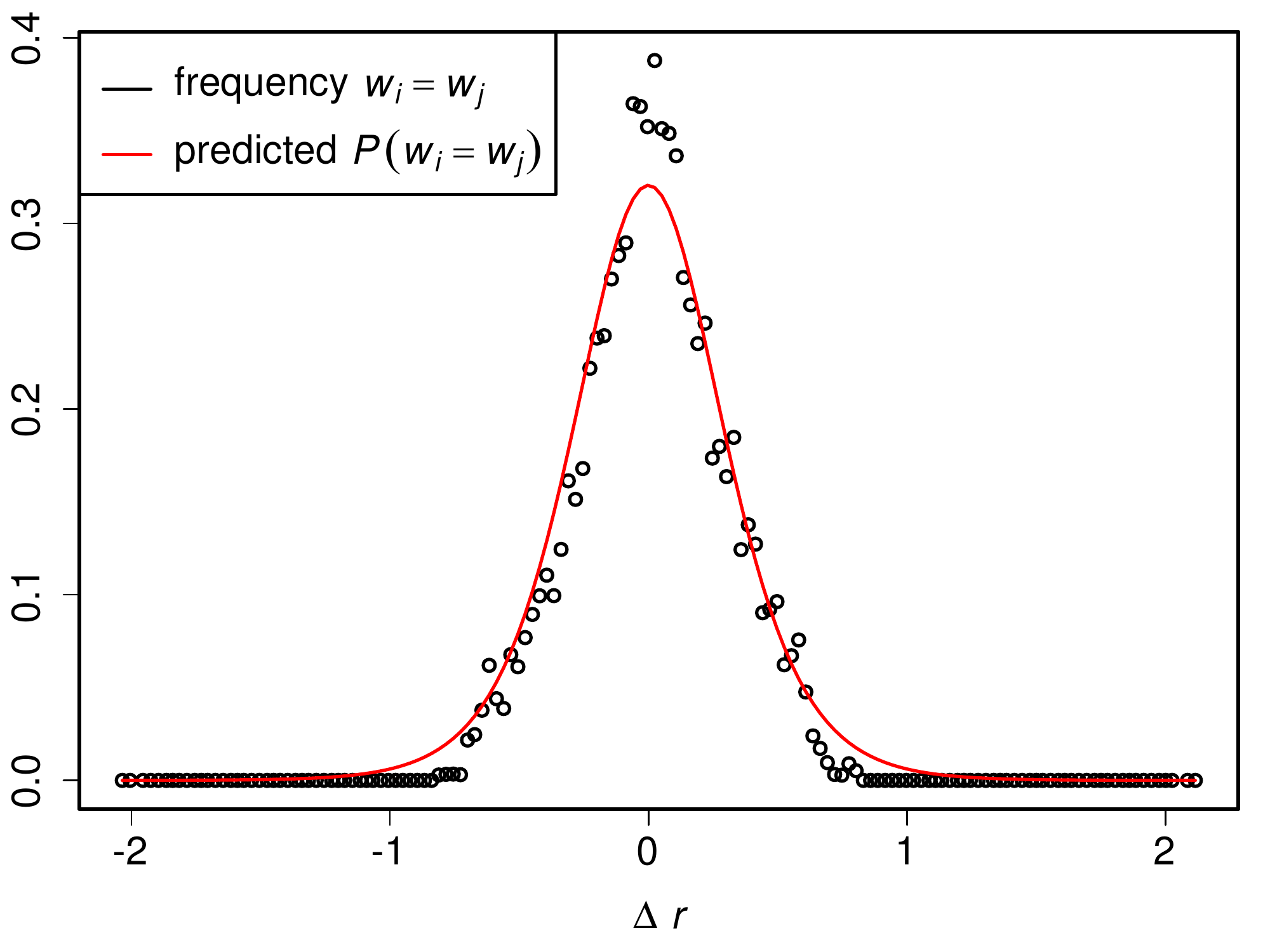}}
  \subfigure[uniform distribution $F$]{\label{fig:gof:unif}\includegraphics[width=0.9\columnwidth]{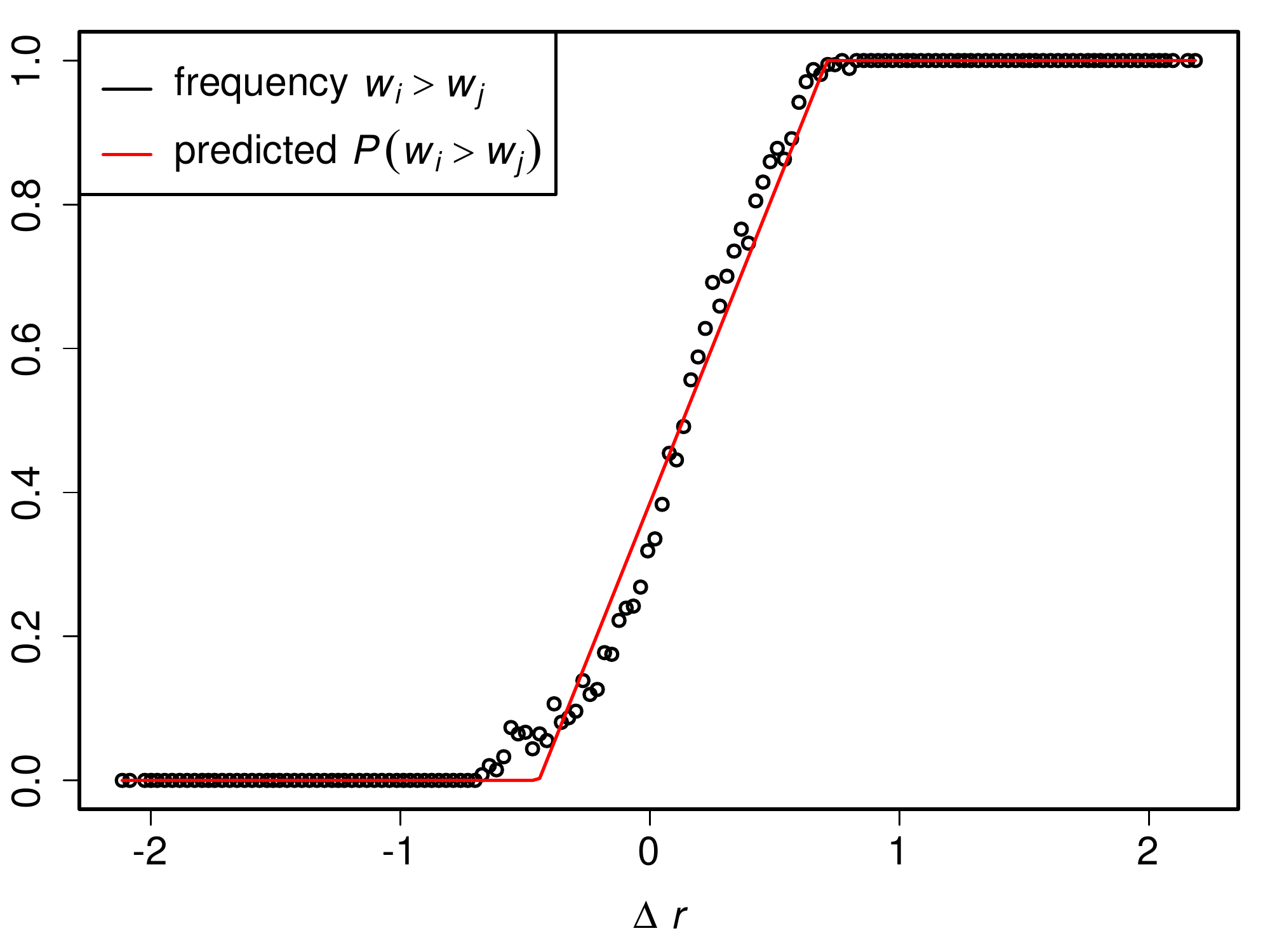}
  \includegraphics[width=0.9\columnwidth]{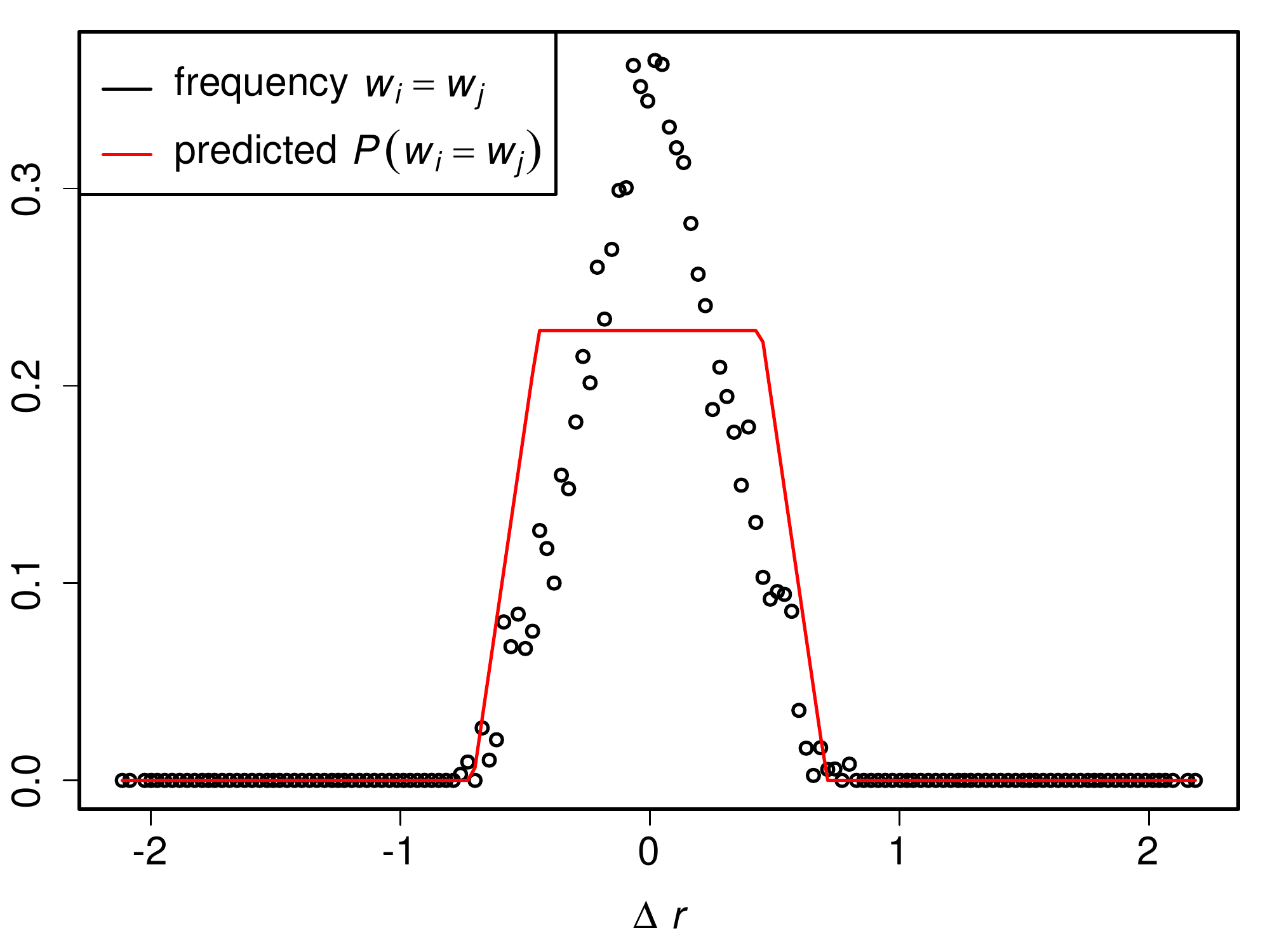}}
  \caption{\label{fig:gof} Comparison of the observed relative frequencies with the probabilities predicted by model (\ref{eq:ansatz}) for 150 bins $\Delta r=r_i-r_j$ and scores estimated by non-linear least squares (LSQ).}
\end{figure*}

\subsection{Goodness of fit}
\label{sec:results:goodness}
Both the ML method and the LSQ method yield
estimators for the sentiment scores $r_i$ even in cases where the model
assumptions in Eq.~(\ref{eq:ansatz}) do not hold. It is thus necessary to
verify that the model explains the experimental observations. Moreover,
it is also interesting to check which of the six solutions (three different
distributions functions in combination with ML or LSQ) provides
the best fit to the experimental data.

A visual way to verify the model consists in cutting the possible score
differences $\Delta r=r_i-r_j$ into bins\footnote{To achieve an equal distribution of positive and negative $\Delta r$, the pair order has been shuffled randomly for the computation of Fig.~\ref{fig:gof}.}
and to compare the observed
frequencies for $w_i>w_j$ and $w_i\approx w_j$ with the probabilities
(\ref{eq:ansatz}) predicted by the model. The resulting plots for the
LSQ estimates are shown in Fig.~\ref{fig:gof}. The figures show that
the model indeed predicts the observed frequencies, albeit better for
wins or losses than for draws. This is no surprise because only one fourth
of the comparisons were draws and consequently the ML or LSQ estimation step
fits the scores rather to the winning probability than to the draw
probability.

\begin{table*}[t]
\begin{center}\begin{small}
\begin{tabular}{l|cccc|cccc}
 \multicolumn{1}{c}{} & \multicolumn{4}{c}{$G=600$} & \multicolumn{4}{c}{$G=800$} \\
  & \multicolumn{2}{c}{LSQ} & \multicolumn{2}{c}{ML} & \multicolumn{2}{c}{LSQ} & \multicolumn{2}{c}{ML} \\
  & $\chi^2$ & $p$ & $\chi^2$ & $p$ & $\chi^2$ & $p$ & $\chi^2$ & $p$\\\hline
normal & $1151$ & $0.001$ & $1286$ & $1.6\cdot 10^{-9}$ & $1450$ & $0.169$ & $1541$ & $0.005$\\
logistic & $1027$ & $0.262$ & $1192$ & $2.2\cdot 10^{-5}$& $1335$ & $0.887$ & $1511$ & $0.019$ \\
uniform & $\infty$ & $0.000$ & $5218$ & $0.000$ & $\infty$ & $0.000$ & $5415$ & $0.000$ \\\hline
$\chi^2_{1-\alpha}$ & \multicolumn{4}{c|}{$1074$} & \multicolumn{4}{c}{1487}
\end{tabular}
\end{small}\end{center}
\caption{\label{tbl:gof} Goodness-of-fit test statistic $\chi^2$ and $p$-values $P(\mbox{\em value}>\chi^2)$ for the different score estimators. The last row gives the corresponding threshold for rejecting the null hypothesis that the model produced the result at a significance level $\alpha=0.05$.}
\end{table*}

For a more formal goodness-of-fit test, we computed the $\chi^2$ statistic
which is the sum of the squared differences between the expected and observed
frequencies. For our paired comparison model, it reads \cite{douglas78}
\begin{align}
\label{eq:chi2}
\chi^2 = \sum_{g=1}^G  & \left(\frac{(W_g - E(W_g))^2}{E(W_g)} \right. +  \\
 & \left. \quad\frac{(D_g - E(D_g))^2}{E(D_g)} + \frac{(L_g - E(L_g))^2}{E(L_g)} \right) \nonumber
\end{align}
where $G$ is the number of bins $I_1,\ldots,I_G$ into which the
$\Delta r=r_i-r_j$ are grouped,
and $W_g$, $D_g$, and $L_g$ are the number of wins, draws, or losses in each
group. The expected number of wins, draws, or losses are computed as the
sum of the respective probabilities according to the model (\ref{eq:ansatz}),
e.g.
\begin{align}
E(W_g) &= \sum_{\Delta r\in I_g} F(\Delta r - t) \quad\mbox{ and }\\
E(D_g) &= \sum_{\Delta r\in I_g} \Big(F(\Delta r + t) - F(\Delta r -t)\Big)
\end{align}
The statistic (\ref{eq:chi2}) is approximately $\chi^2$ distributed when the
number of members in each group is not too small. The number of degrees of
freedom is $2G-n-2$, because out of the three possible outcomes in each group only
two are a free choice (when the outcome is neither win nor draw, it must be a
loss) and there are $n+1$ fitted parameters in the model: the $n$ ratings 
$r_i$ and the draw width $t$.

Douglas suggested in \cite{douglas78} to use each
possible score difference as its own group, but this would mean that we would
only have two items per group and the test statistic would not be $\chi^2$
distributed. We therefore  formed the groups
as quantiles among the $|\Delta r|$ values\footnote{We have computed the quantiles on basis of $|\Delta r|$ instead of $\Delta r$ because otherwise the value of the test statistic would depend on the pair order.}
such that each group had
approximately the same number of samples, a method commonly deployed
in testing logistic regressions \cite{hosmer80}.

The resulting $\chi^2$ values are listed in Table \ref{tbl:gof}.
Surprisingly, the LSQ estimators lead to a clearly better goodness-of-fit
than the ML estimators, with the ML estimators even yielding such a high
value for $\chi^2$ that the model would be rejected as unlikely.
As usual for $\chi^2$ tests, the result varies with the number of
groups however, and the $p$-values are different for different
numbers of groups $G$. Nevertheless, the {\em relative} values of $\chi^2$
for fixed $G$ allow to compare different estimation methods, and in all cases
the LSQ estimator yielded a clearly better fit. With respect to the
distribution function $F$, the uniform distribution provides an impossible
LSQ fit
because it leads to wins with zero probability in some groups. $\chi^2$ is
smallest for the logistic choice for $F$, which is therefore the model which
conforms best to the observed test person answers.

We conclude that, for estimating the sentiment scores, the non-linear
LSQ estimators
given by Eqs.~(\ref{eq:ss}) and (\ref{eq:t-lsq}) should be used.


\subsection{Score values}
\label{sec:results:scores}

\begin{figure}[t]
  \centering
  \includegraphics[width=0.9\columnwidth]{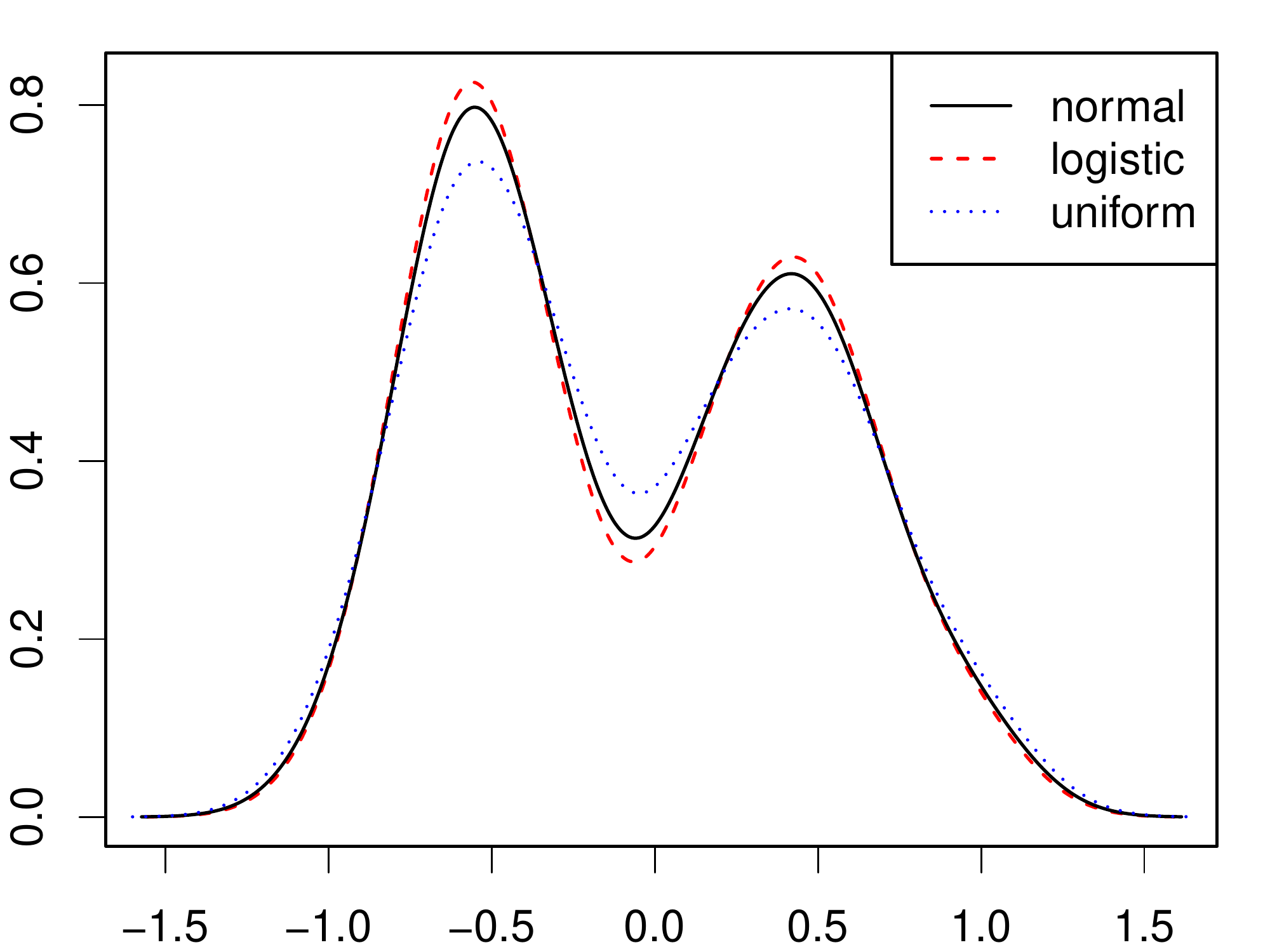}
  \caption{\label{fig:density} Kernel density plot of the polarity score
    distribution in our sentiment lexicon, estimated with non-linear LSQ for
    different cumulative distribution functions $F$.}
\end{figure}

\begin{table}[b]
\begin{center}\begin{small}
\begin{tabular}{l|rrr}
   & {\em normal} & {\em logistic} & {\em uniform} \\\hline
unpraktisch ({\em unpractical})   & -0.378 & -0.396 & -0.344 \\
r\"ude  ({\em uncouth})           & -0.628 & -0.629 & -0.631 \\
t  ({\em draw width})             &  0.126 &  0.122 &  0.132 \\\hline
$P(\mbox{unpraktisch} > \mbox{r\"ude})$ & 0.644 & 0.647 & 0.635 \\
$P(\mbox{unpraktisch} \approx \mbox{r\"ude})$ & 0.226 & 0.227 & 0.228 \\
\end{tabular}
\end{small}\end{center}
\caption{\label{tbl:Fscores} Comparison of scores obtained for different distribution functions $F$.}
\end{table}

For the 199 different words, we estimated the polarity scores with the
LSQ method with the three distribution functions of Fig.~\ref{fig:cdf}.
The draw width $t$ turned out to be 0.126 for the normal distribution, 0.122
for the logistic distribution, and 0.132 for the uniform distribution.
Fig.~\ref{fig:density} shows a kernel density plot \cite{sheather91} for the
resulting score distributions. The valley around zero (neutrality) is due
to the fact that the words were drawn from the SentiWS data, which only
contains positive or negative words. The comparative shapes are as expected
from Fig.~\ref{fig:cdf}: the slightly steeper slope of the logistic
distribution $F(x)$ at $x=0$ leads to a slightly stronger separation
of the positive and negative scores. The range of the
score values depends on the distribution $F$, too: for the normal distribution
it is $[-1.058,1.099]$, for the logistic distribution $[-1.042,1.083]$, and for
the uniform distribution $[-1.080,1.118]$. 
The example in Table \ref{tbl:Fscores} shows that the
varying difference in ratings for the different distribution functions
has only little effect on the ``winning probabilities'', because the greater
rating difference for the uniform distribution, e.g., is compensated by its
greater draw width $t$.

\begin{table}[t]
\begin{center}\begin{small}
\begin{tabular}{l|r|rr}
  {\em adjective} & \multicolumn{1}{c|}{$r_{\mbox{\scriptsize\em direct}}$} & \multicolumn{1}{c}{$r_{\mbox{\scriptsize\em paired}}$} & \multicolumn{1}{c}{$\sigma_{\mbox{\scriptsize\it boot}}$} \\\hline
paradiesisch ({\em paradisaical})  &  1.00 &  0.966 & 0.039 \\
wunderbar ({\em wonderful)}       &  1.00 &  0.915 & 0.045 \\
perfekt  ({\em perfect})          &  0.95 &  1.083 & 0.057 \\
traumhaft ({\em dreamlike})       &  0.95 &  1.041 & 0.050 \\
prima ({\em great)}               &  0.75 &  0.793 & 0.041 \\
zufrieden ({\em contented})       &  0.75 &  0.613 & 0.034 \\ 
kinderleicht ({\em easy-peasy})  &  0.50 &  0.462 & 0.034 \\
lebensf\"ahig ({\em viable})      &  0.50 &  0.351 & 0.032 \\
ausgeweitet ({\em expanded})      &  0.05 & -0.012 & 0.027 \\
verbindlich ({\em binding})        &  0.00 &  0.143 & 0.023 \\
kontrovers ({\em controversial})  & -0.05 & -0.268 & 0.029 \\
unpraktisch ({\em unpractical})   & -0.50 & -0.396 & 0.026 \\
r\"ude  ({\em uncouth})           & -0.50 & -0.629 & 0.026 \\ 
falsch  ({\em wrong})             & -0.75 & -0.627 & 0.025 \\
unbarmherzig ({\em merciless})    & -0.75 & -0.772 & 0.027 \\ 
erb\"armlich ({\em wretched})     & -1.00 & -0.804 & 0.025 \\
t\"odlich ({\em deadly})          & -1.00 & -1.042 & 0.031 \\
\end{tabular}
\end{small}\end{center}
\caption{\label{tbl:examplescores} Example scores from average direct assignment and
  paired comparisons with the logistic distribution.}
\end{table}

It is interesting to compare the scores from paired comparisons for words
which have obtained the same score from direct assignment on the five
grade scale. The examples in table \ref{tbl:examplescores} show that
the paired comparisons indeed lead to a different and finer rating scheme
than averaging over coarse polarity scores from direct assignments, and that
they even can lead to a reversed rank order (see, e.g., ``traumhaft'' and
``wunderbar''). We have also estimated
the variances of the polarity score estimates as the bootstrap variance
$\sigma_{\mbox{\scriptsize\it boot}}^2$ from 200 bootstrap replications
of all paired results \cite{efron86}. These can be used to
test whether, for $r_i\neq r_j$, the score difference is significant by computing
the $p$-value $1-\Phi\left(|r_i-r_j|/\sqrt{\sigma_i^2 + \sigma_j^2}\right)$,
where $\Phi$ is the distribution
function of the standard normal distribution. For the words ``unpraktisch''
and ``r\"ude'', e.g., the $p$-value is much less than 5\%
and the difference is therefore statistically significant, although they
obtained identical scores on the five grade scale from direct estimation.

\subsection{Adding new words}
\label{sec:results:adding}
To obtain a lower bound for the error in estimating scores for unknown words,
we have first computed the scores for all words with the
estimators for one unknown rating $r$ as described in section
\ref{sec:method:case1}, where each word was compared with all other words and
the scores $q_i$ for other words were considered
to be known from the results in the preceding section.
For the logistic distribution, the mean absolute error with respect to the
non-linear LSQ score was much higher for the maximum-likelihood estimator
($0.012$) than for method-of-moments estimator ($<10^{-5}$)
computed with Eq.~(\ref{eq:W+D/2approx}). 
This was to be expected because the ``ground truth score'' was also
based on the difference between observed and expected wins and draws.
With respect to the ML score, the ML estimator was 
better ($0.002$ versus $0.012$), but, as we have seen in section
\ref{sec:results:goodness}, the ML scores are a much poorer model fit and
therefore should not be used as a point of reference. We therefore have used
the method-of-moments estimator (\ref{eq:W+D/2approx}) in our evaluations.

\begin{figure}[t]
  \centering
  \includegraphics[width=1.0\columnwidth]{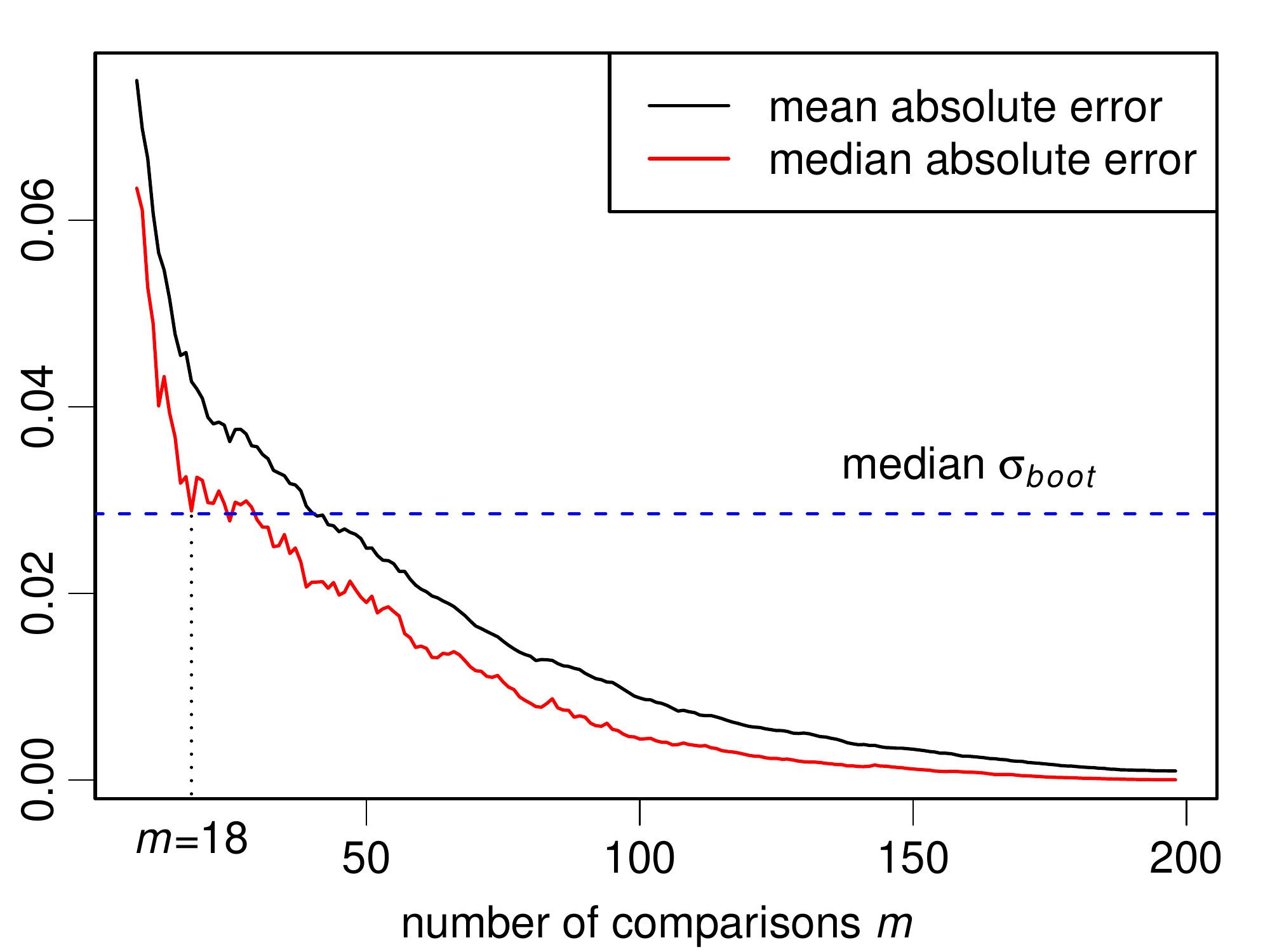}
  \caption{\label{fig:results-m} Mean and median absolute error from
    leave-one-out as a function of the number of comparisons
    in Algorithm \ref{alg:nw}.}
\end{figure}

For a reasonable recommendation for the number of incomplete comparisons, we
have varied the number of comparisons $m$ in a leave-one-out two-fold
application of Algorithm \ref{alg:nw} on basis of the two-fold round robin
experiment.
The results are shown in Fig.~\ref{fig:results-m}. As a point of reference,
the median of the bootstrap standard deviation 
$\sigma_{\mbox{\scriptsize\it boot}}$ of the ground truth scores is given too
($0.029$). The median of the absolute
score errors reaches this reference value first for $m=18$, which corresponds
to picking $m-\log_2(n)\approx 10$ neighbors after Silverstein \& Farrel's
method.

To check whether our word selection method based on
Silverstein \& Farrel's method actually is an improvement over picking words
for comparison at random, we did a 100-fold Monte-Carlo experiment with
choosing $18$ words at random and computing from all corresponding
two-fold comparisons the method-of-moments score estimate. This yielded on
average a mean error of $0.067$ and a median error $0.053$, which are
considerably greater than the errors of Algorithm \ref{alg:nw}.
We therefore conclude that incomplete comparisons
with only 18 out of 199 words yield a reasonably accurate score estimate,
provided the words are selected with our method.


\subsection{Comparison to corpus-based lexica}
\label{sec:results:comparison}
The polarity scores computed in our experiments provide nice ground
truth data for the evaluation of corpus-based polarity scores. We therefore
compared the scores from SentiWS 1.8, SenticNet 3.0, SenticNet 4.0, and
SentiWordNet 3.0 with the scores computed from test persons' paired
comparisons. SenticNet and SentiWordNet only contain English words,
from which we have computed scores for the German words by translating each
German word with both of the German-English dictionaries from
{\em www.dict.cc} and {\em www.freedict.org} and by averaging
the corresponding scores, a method that we call ``averaged translation''.

A natural measure for the closeness between lists of polarity scores is
Pearson's correlation coefficient $r_p$, which has the advantage
that it is invariant both under scale and translation of the variables.
This is crucial in our case, because score values from paired comparisons
allow for arbitrary shift and scale as explained in section \ref{sec:method}.
$r_p$ is highest for a linear relationship
and smaller for other monotonous relationships. As can be seen in Table
\ref{tbl:correlation}, this means that its value
depends on the shape of the model distribution function $F$, albeit only
slightly. Whatever function
is used, the correlation between the scores from direct assignment and
paired comparison is very strong. This was to be expected, because both values
stem from the same test persons.

\begin{table}[t]
\begin{center}\begin{small}
\begin{tabular}{l|ccc}
  & \multicolumn{3}{c}{\em choice for $F$} \\
  & \multicolumn{1}{c}{\em normal} & \multicolumn{1}{c}{\em logistic} & \multicolumn{1}{c}{\em uniform} \\\hline
{\em direct}    &  $0.977$ &  $0.978$ & $0.973$ \\
{\em SentiWS}    &  $0.714$ &  $0.715$ & $0.713$ \\
{\em SenticNet 3.0}    &  $0.759$ &  $0.763$ & $0.752$ \\
{\em SenticNet 4.0}    &  \boldmath$0.824$ &  \boldmath$0.828$ & \boldmath$0.817$ \\
{\em SentiWordNet 3.0}    &  $0.792$ &  $0.794$ & $0.789$ \\
\end{tabular}
\end{small}\end{center}
\caption{\label{tbl:correlation} Pearson Correlation $r_p$ of the parity scores from the paired comparison with that of direct assignment and corpus-based methods.}
\end{table}

\begin{figure*}[t]
  \centering
  \subfigure[SentiWS]{\label{fig:scatter:sentiws}\includegraphics[width=0.49\columnwidth]{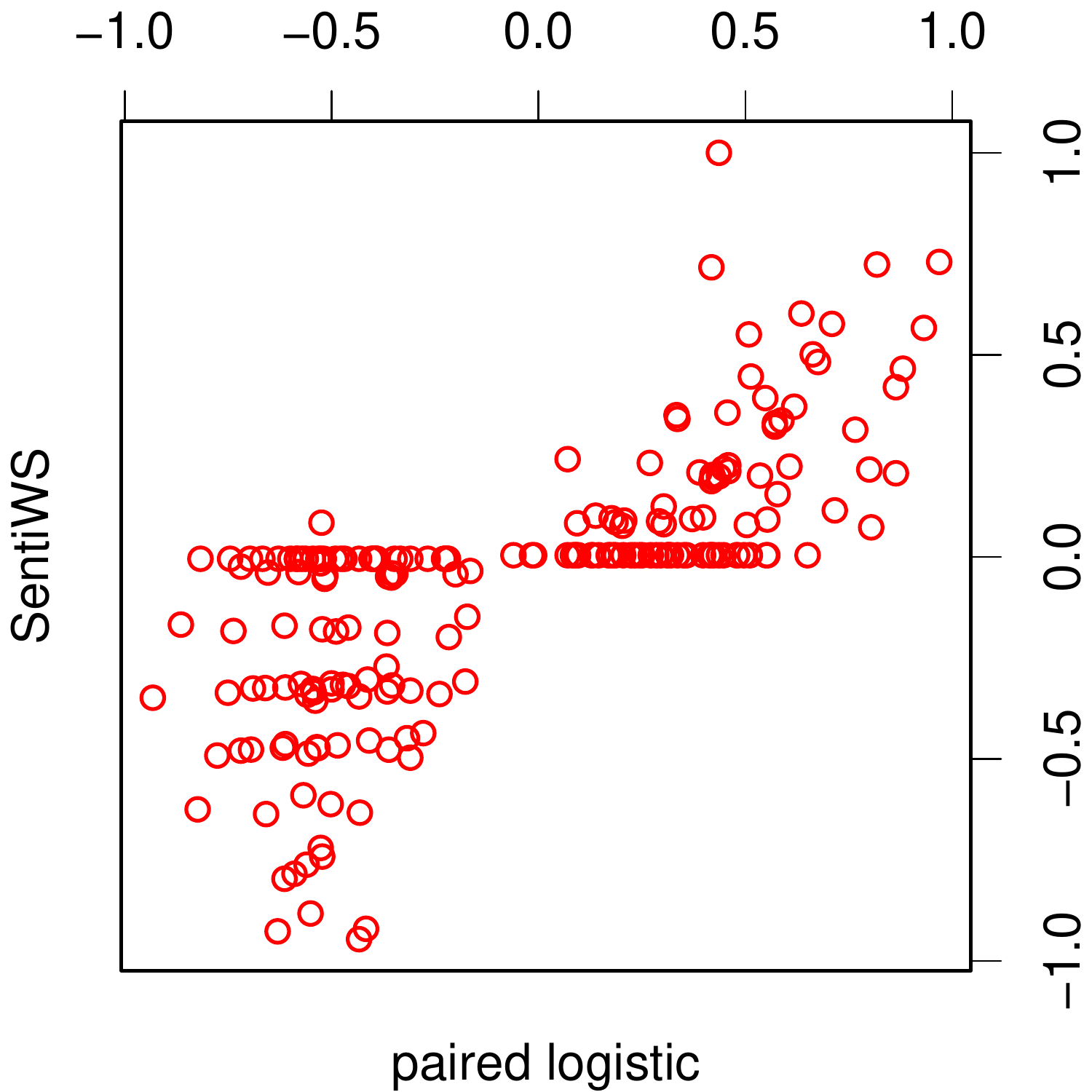}}
  \subfigure[SenticNet 3.0]{\label{fig:scatter:senticnet3}\includegraphics[width=0.49\columnwidth]{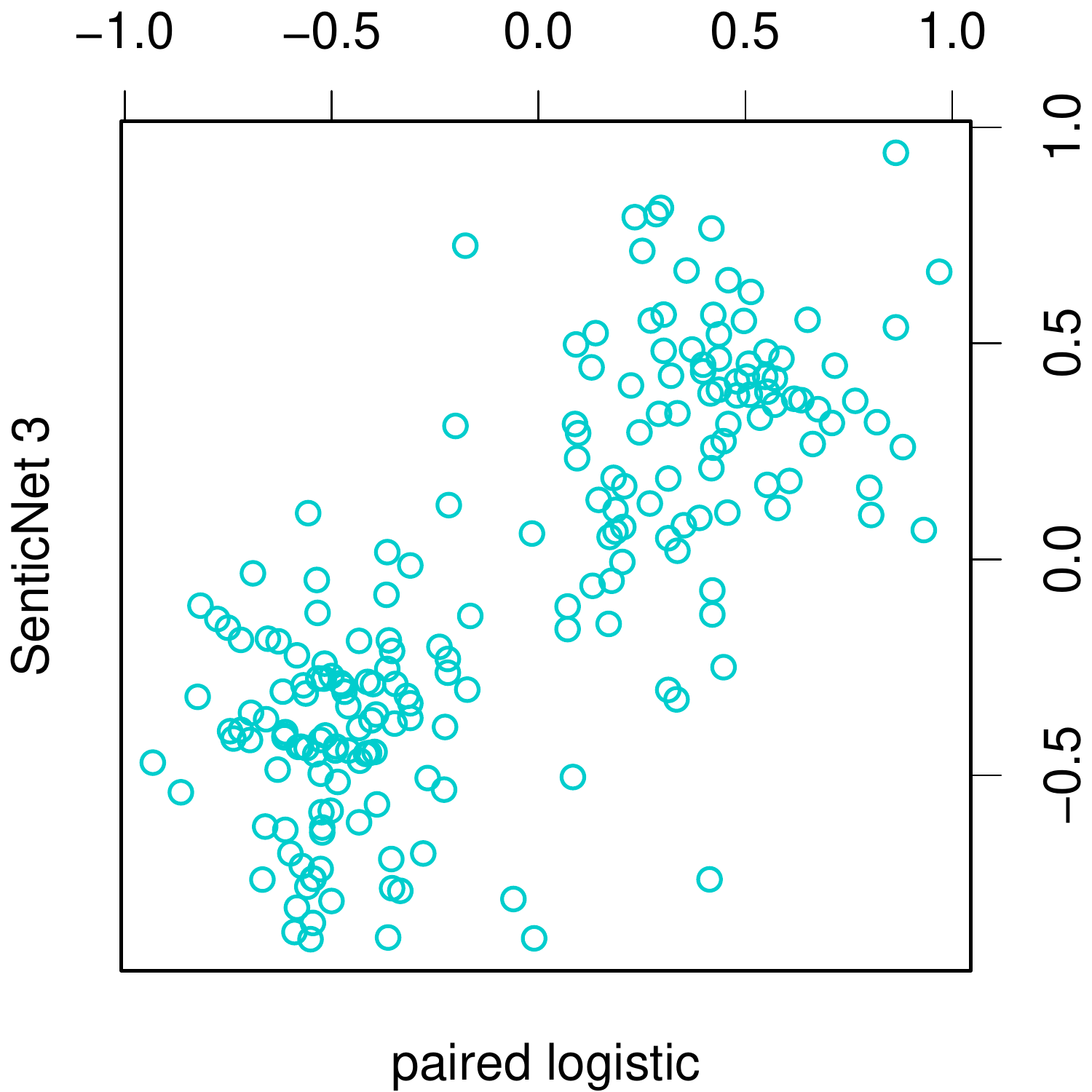}}
  \subfigure[SenticNet 4.0]{\label{fig:scatter:senticnet4}\includegraphics[width=0.49\columnwidth]{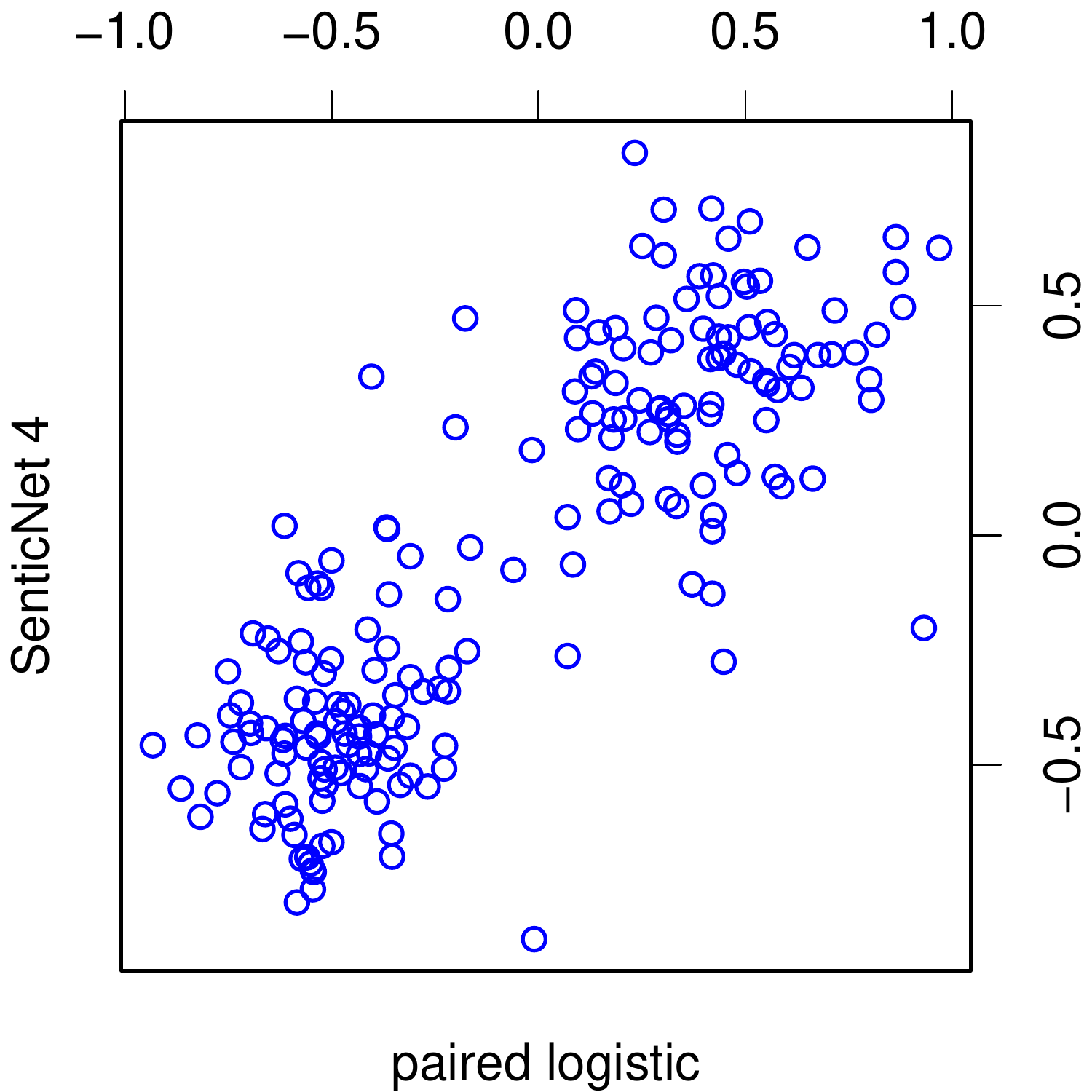}}
  \subfigure[SentiWordNet 3.0]{\label{fig:scatter:sentiwordnet}\includegraphics[width=0.49\columnwidth]{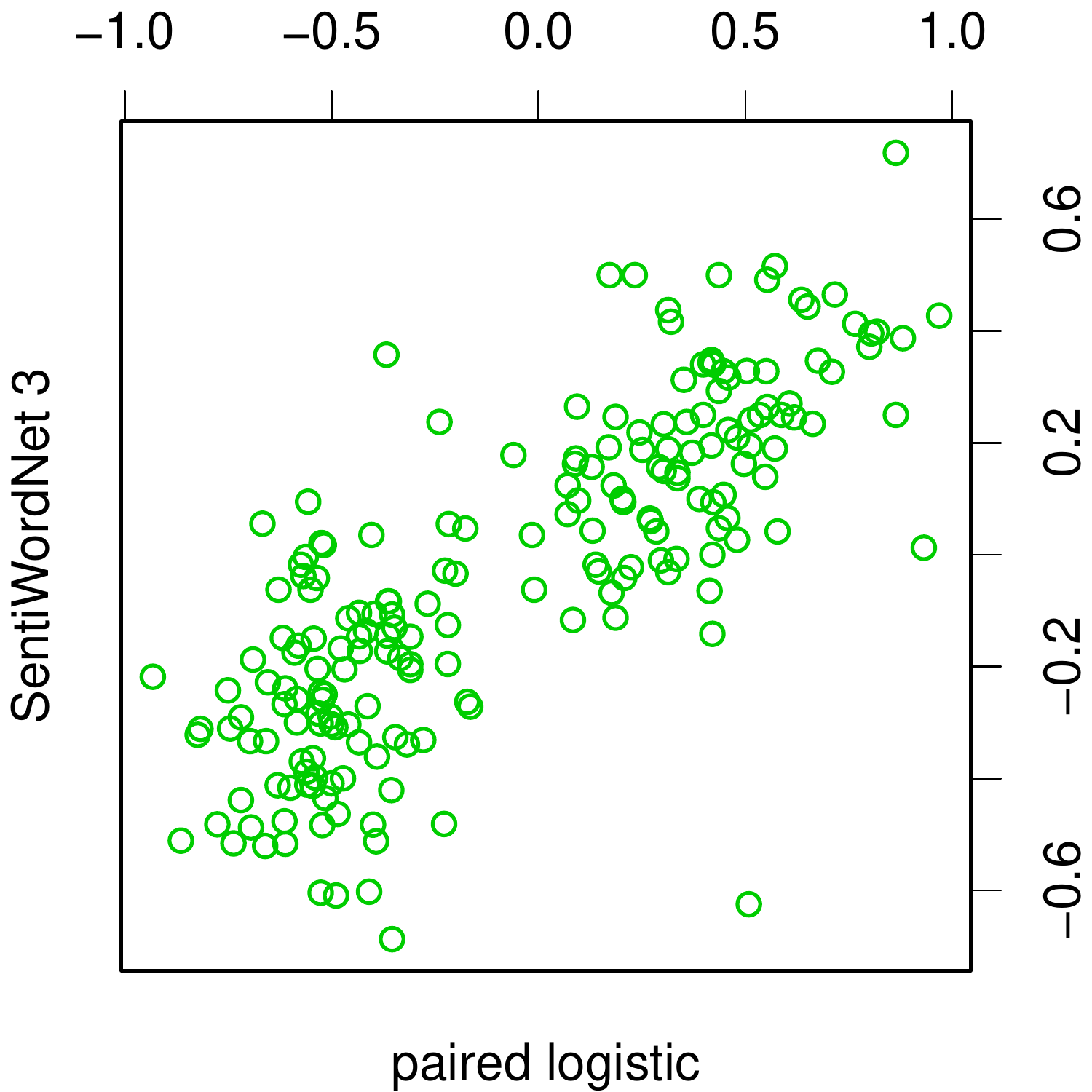}}
  \caption{\label{fig:scatter} Scatter plots comparing corpus-based scores with scores from paired comparisons.}
\end{figure*}

From the corpus based lexica, the correlation with the paired scores is
highest for SenticNet 4.0.
According to the significance tests in the R package {\em cocor}
\cite{diedenhofen15}, the difference to the correlation of SentiWordNet is
not statistically significant at a 5\% significance level, but it is 
significant with respect
to the other two corpus based lexica, including SenticNet 3.0.

\begin{figure}[t]
  \centering
  \includegraphics[width=0.9\columnwidth]{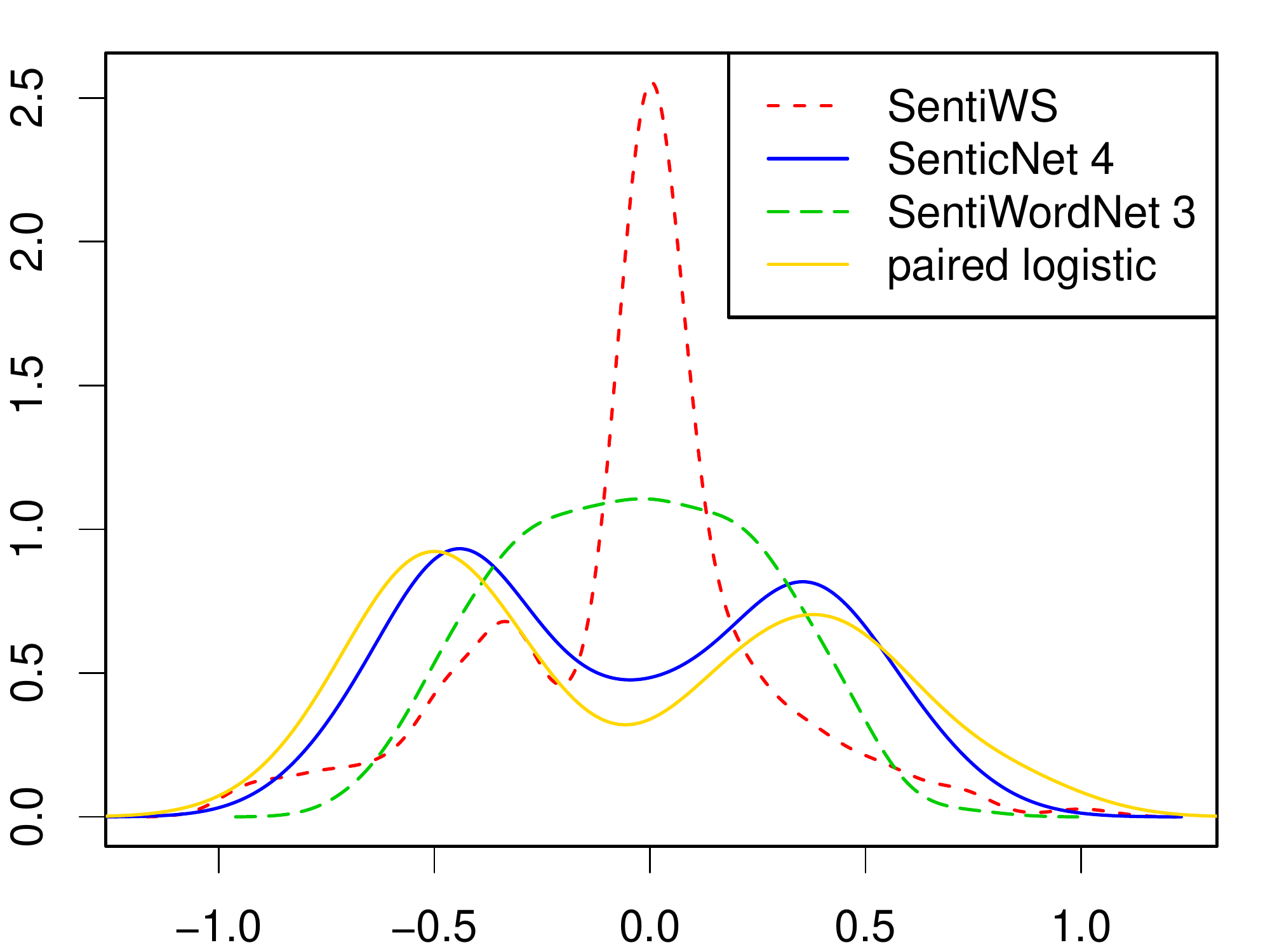}
  \caption{\label{fig:comparedensity} Kernel density plots of the polarity
    score distributions. The paired comparison scores have been scaled to
    the range $[-1,1]$.}
\end{figure}

Another comparison criterion that is in favor of SenticNet 4.0 can be seen
in Fig.~\ref{fig:comparedensity}: the bimodal shape of the score distribution
is only reflected in the SenticNet scores. SentiWordNet has an unimodal
score distribution centered around zero, and SentiWS has many identical scores
with values $0.0040$ and $-0.0048$, which show up as horizontal lines
in Fig.~\ref{fig:scatter:sentiws}. This peculiar distribution of the
SentiWS scores was also reported in the original paper presenting the
SentiWS data set by Remus et al.~(see Fig.~1 in \cite{remus10}).
The identical scores show up in Fig.~\ref{fig:comparedensity}
as a peak around neutrality, which corresponds to a {\em valley} (sic!)
in the score distribution from paired comparisons.

\pagebreak[4]
As an alternative to our ``averaged translation'' approach, the third party Python module {\em senticnetapi}\footnote{\url{https://github.com/yurimalheiros/senticnetapi}}
also provides sentiment scores for German words, which cannot be recommended,
however. These have been obtained from
SenticNet 4.0 by automatic translation from English
to German, which is the reverse direction than our approach. This has the effect
that only 35\% of the words in our lexicon have matches in senticnetapi,
and these only have a correlation of $0.694$, which is even less than the value
for SentiWS.

Based
on these observations, we consider the polarity scores from SenticNet 4.0
the most reliable among the investigated corpus based lexica. For foreign
languages, we recommend to translate from the foreign language to English and
compute the mean score among the matches, rather than to translate from English
into the foreign language.

\section{Conclusions}
\label{sec:conclusions}
The new sentiment lexicon from paired comparison is a useful resource that
can be used for different aims. It can be used, e.g., as ground truth data
for testing and comparing automatic corpus-based methods for building
sentiment lexica, as we did in section \ref{sec:results:comparison}.
Our results suggest that the most reliable corpus-based sentiment lexicon,
among the tested ones, is SenticNet 4.0. When it is used with non-English
languages, the ``averaged translation'' method should be used rather than
a translation from English into the foreign language.

Our lexicon can also be used as a starting point for building specialized lexica
for polarity studies. The method for adding new words makes the method
of paired comparison applicable to studies with an arbitrary vocabulary
because it yields accurate polarity scores even for rare words. We make
our new lexicon, a GUI for adding words via test users, and the software
for computing scores freely available on our website\footnote{\url{http://informatik.hsnr.de/~dalitz/data/sentimentlexicon/}}.

Although our results suggest that only $18$ two-fold comparisons are 
sufficient for estimating the score of a new, yet unknown word, this
is still very time-consuming in practice. It would thus be interesting to
combine the paired-comparison method with a corpus-based lexicon, such that
only missing words or words estimated with a low confidence are added
by experiments with test persons. Such a hybrid approach would be an
interesting subject for further research.

\bibliographystyle{ieeetr}
\bibliography{paired-comparison}

\end{document}